# The Exploitation of Distance Distributions for Clustering

Michael C. Thrun

*Databionics Research Group, Philipps-University of Marburg, D-35032 Marburg, Germany*
*mthrun@mathematik.uni-marburg.de*

Although distance measures are used in many machine learning algorithms, the literature on the context-independent selection and evaluation of distance measures is limited in the sense that prior knowledge is used. In cluster analysis, current studies evaluate the choice of distance measure after applying unsupervised methods based on error probabilities, implicitly setting the goal of reproducing predefined partitions in data. Such studies use clusters of data that are often based on the context of the data as well as the custom goal of the specific study. Depending on the data context, different properties for distance distributions are judged to be relevant for appropriate distance selection. However, if cluster analysis is based on the task of finding similar partitions of data, then the intrapartition distances should be smaller than the interpartition distances. By systematically investigating this specification using distribution analysis through a mirrored-density plot, it is shown that multimodal distance distributions are preferable in cluster analysis. As a consequence, it is advantageous to model distance distributions with Gaussian mixtures prior to the evaluation phase of unsupervised methods. Experiments are performed on several artificial datasets and natural datasets for the task of clustering.

## 1. Introduction

Entire classes of clustering techniques rely on distances [Fayyad et al., 1996; Tan et al., 2006; Yang et al., 2019]. Usually, popular algorithms use either Euclidean distances [Yang et al., 2019], because there is no theory to select an appropriate distance measure [Brazma/Vilo, 2000], or custom distances that are closely related to the problem (e.g., [Hampapur/Bolle, 2001; Li et al., 2004; Cha, 2007; Yujian/Bo, 2007; Kumar/Vassilvitskii, 2010; Cao et al., 2013]). Sometimes, the choice of the distance measure can be more important than the algorithm itself [Gentleman et al., 2005; Priness et al., 2007] and can improve the quality of supervised (e.g.,[Phyu, 2009]) and unsupervised methods (e.g., [Priness et al., 2007]).

The most often used Euclidean distance measure requires accounting for correlations (e.g., [Cormack, 1971; Mimmack et al., 2001]) and either normalizing the data to avoid an undesired emphasis on features with large ranges and variances [Mörchen, 2006] (see also [Jain/Dubes, 1988; Weihs/Szepannek, 2009]) or weighting the distance [Mitchell, 1997]. More generally, for commonly used $L_k$ norms, the concept of proximity in high dimensions is sensitive to the value of k, and the Euclidean $L_2$ norm is not well suited for clustering in high-dimensional space [Aggarwal et al., 2001]. To account for the above disadvantages of the Euclidean distance, the selection of a more appropriate distance measure can be an important step in modeling data, but it often remains neglected.

Apart from learning distance measures for specific problems (e.g., [Xing et al., 2003]), the choice of a distance metric is usually based on either statistical properties or distributions. In the first case, domain experts provide specific examples of what they consider similar [Xing et al., 2003]. Such examples do not always satisfy the triangle inequality. In the second case, distance distributions are compared in spatial analytical modeling based on descriptive statistics with the implicit assumption of a unimodal distribution [Shahid et al., 2009]. For genome-wide datasets, it has been concluded that an appropriate distance measure should have extreme values of kurtosis and skewness [Glazko et al., 2005]. As a third alternative, distance distributions with two peaks of a specific distance measure for a specific application have been reported, but this effect has not been investigated thoroughly (e.g., [Bozkaya/Ozsoyoglu, 1997; Beyer et al., 1999a; De Smet et al., 2002; Cao et al., 2013; Thrun, 2019a]). Since the work of Basseville [Basseville, 1989], the main performance criterion for distances still remains the evaluation of error probabilities through the comparison of specific classification problems [Sooful/Botha, 2002; Gavin et al., 2003; Schenker et al., 2003; Finch, 2005; Bharkad/Kokare, 2011; Wang et al., 2013; Yang et al., 2019].

All these studies have the common goal of reproducing prior classifications. As a consequence, these cluster structures are predefined within a specific semantic context because the definition of a cluster remains a matter of ongoing discussion[Bonner, 1964; Hennig et al., 2015]. It is not investigated whether these datasets possess structures based on distances or whether the prior classification is only a result of one out of several possible optimal solutions [Färber et al., 2010]. This makes the generalization of these studies challenging. Furthermore, the curse of dimensionality [Verleysen et al., 2003] is ignored: as the dimension increases, the distance between any two points tends towards a constant.

In this work, a data-driven approach for distance selection is proposed based on a systematic investigation of distance distributions. If the goal of cluster analysis is specified for the detection of similar partitions of data, this restriction can be used to select a distance for cluster analysis and evaluate a clustering method for data based on Bayesian decision boundaries. In section II, similar partitions of data will be defined as distance-based cluster structures. As an example, Fig. 1 presents two scatter plots of two datasets, one with distance-based structures for two clusters and one without distance-based structures.

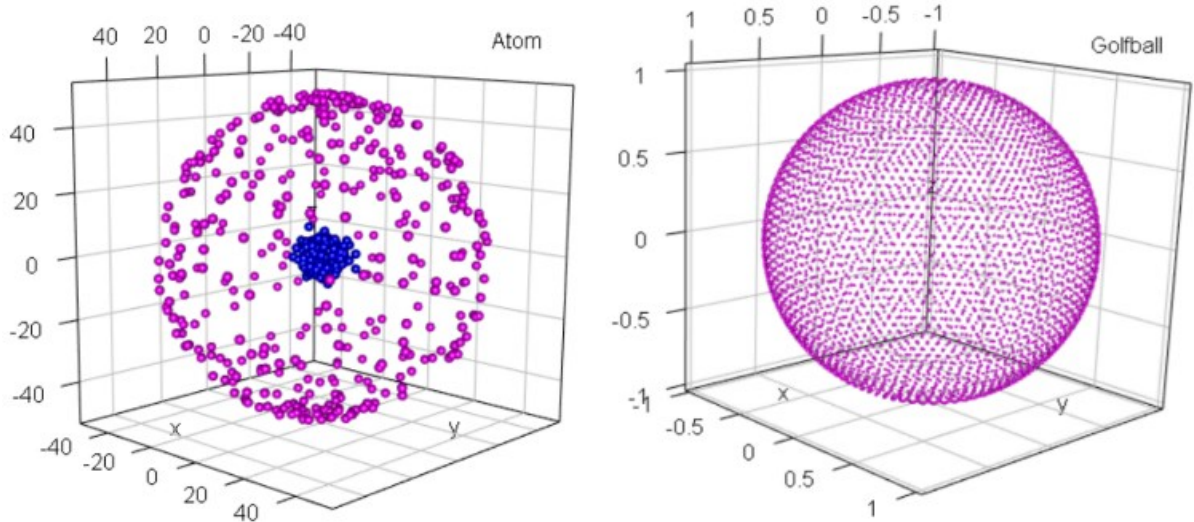

Fig. 1. Two clusters in 3D are shown on the left, and an empty sphere without distance-based cluster structures is shown on the right. The predefined classification is indicated by color, and the datasets of arbitrary sample size are published in [Thrun/Ultsch, 2020a].

In sum, the literature provides conflicting and context-based guidelines for how to select distance metrics based on distribution properties. The contribution is to show that if the goal of cluster analysis is restricted to extracting information based on relative relationships between data points, then distance selection based on a multimodal distance distribution is preferable. For cases in which such so-called distance-based structures are relevant, this work proposes exploiting mirrored-density plots [Thrun et al., 2020] as a tool for quickly identifying multimodality in distance distributions. Then, it is advantageous to model the distance distribution with Gaussian mixture models because such models enable the application of Bayes' theorem, with which specific Bayesian boundaries can be identified in the distance distribution. Combining Bayesian boundaries with intrapartition distance distributions allows the investigation of whether distance-based clustering algorithms yield appropriate clusterings. Contrary to prior works, datasets are used for which natural distance-based cluster structures were identified previously [Thrun/Ultsch, 2020a].

The article is structured as follows: The methods section introduces the theoretical concept of cluster analysis within a clear mathematical framework that allows for distance selection as well as the definitions required for modeling distance distributions. The compelling consequence of the framework is outlined on two artificial distance-based datasets, which show that the relevant property of distance distributions is multimodality. Using two natural examples with distance-based structures [Thrun/Ultsch, 2020a], it is shown how the property of multimodality can be exploited with Gaussian mixture models and that clusterings can be evaluated with intrapartition and interpartition distance distributions. After a discussion, the article ends with a conclusion.

## 2. Methods

The task of clustering or partitioning can be described as the search for clusters that have similar objects inside them [Bouveyron et al., 2012] and dissimilar objects outside w.r.t. each partition. In such cases, often a distance measure can be defined so that the triangle inequality holds. Then, the distance measure induces a metric space for the data and can define distance-based structures. Satisfying metric properties are desirable because the "metric-associated benefits include well-defined point neighborhoods, advanced indexing through metric trees, provable convergence, guarantees for embedding, and intuitive result interpretation"[Yang et al., 2019]. Consequently, the relevant information for cluster analysis lies in the relative relationships between high-dimensional data points. The implicit assumption is that the intrapartition distances (intra-pds) are smaller within a cluster than the interpartition distances (inter-pds) between clusters of data. In cluster analysis, clusters with this definition are often called natural clusters [Duda et al., 2001] or well-separated clusters[Tan et al., 2006].

In datasets, such cluster structures are called distance-based, and it is a complex task to decide whether a dataset possesses cluster structures. Typically, this can be decided after the clustering process by using either evaluation metrics[Handl et al., 2005; Arbelaitz et al., 2013] or visualization of the clustering results using projection methods [Venna et al., 2010; Ultsch/Thrun, 2017; Thrun, 2018]. If the results do not show distance-based structures, then the user has the trying task of deciding whether the distance measure used is incorrect, some of the preprocessing steps are incorrect, or the data itself does not possess any distance-based structure.

By exploiting the theoretical concept below, the existence of distance-based structures and the choice of distance can be explored by distribution analysis. For these purposes, a data-driven selection of a distance measure independent of the underlying context of data is presented. The theoretical concept behind the approach is based on the two following assumptions:

Let I be a finite subset of a metric space with a distance function $d(l,j)$; then, the matrix $D\big(d(l,j)\big)_{l,j\in I}$ is called a distance matrix of I (c.f.[Neumaier, 1981]). The distance matrix $D_{ij}$ satisfies the four conditions that the diagonal entries are all zero ($D_{ii} = 0, \forall\ 1 \leq i \leq N$), the other entries are positive ($D_{ij} > 0, \forall\ i \neq j$) and symmetric ($D_{ij} = D_{ji}$) and for any $i, j, D_{ij} \leq D_{ik} + D_{kj}, \forall k$ the triangle inequality holds. It follows that a distance matrix has to be symmetric with nonnegative elements[Gower/Legendre, 1986]. Using this, the distance feature $df$ is defined as the vector with the elements of the upper triangle of the distance matrix.

An overview of specific distance measures for specific applications can be found in [Deza/Deza, 2009]. The **first assumption** states that a natural partition of data is described as clusters of similar objects [Bouveyron et al., 2012], where a distance measure defines (dis-)similarity. The distance between two similar points $l, j \in I$ is small, whereas that between two dissimilar points $l, j \in I$ is large. Then, the choice of distance metric can be guided by the goal of finding natural partitions in data defined as distance-based cluster structures (see[Bock, 1974] for details about transformations from dissimilarity to distance).

## 2.1. Intrapartition distances

Let $t_p \subset I$ and $t_q \subset I$ be two distinct partitions such that $\forall\ p, q \in \{1, \dots, k\}$ and $p \neq q$, $t_p \cap t_q = \emptyset$; then, the distance $Intra(t_p) \coloneqq D_{l,j}$ between two data points $j, l \in t_p$, is called the intrapartition distance (intra-pd). Then, $df_{intra(t_p)}$ is the feature that contains all intrapartition distances of partition $t_p$ and $f_{intra(t_p)}$ the PDF of the distribution of that feature and partition.

## 2.2. Interpartition distances

Let $t_p \subset I$ and $t_q \subset I$ be two partitions such that $p, q \in \{1, \dots, k\}$, $t_p \cap t_q = \emptyset$, and $p \neq q$; then, the distance $Inter(t_p, t_q) = D_{l,j}$ between two data points $j$ and $l$ in the two partitions, $j \in t_p$ and $l \in t_q$, is called the interpartition distance (inter-pd).

## 2.3. Probability density distribution PDF

The distribution of the distance feature $df$ is regarded as an approximation of its probability density function (PDF). The empirical estimation of the PDF is a difficult task to perform. Hence, Pareto density estimation (PDE) [Ultsch, 2005] is combined here with the mirrored density plot (MD plot) [Thrun et al., 2020]. PDE is particularly suitable for the discovery of structures in continuous data [Ultsch, 2005]. The MD plot is a schematic plot enabling the visualization of several features at once, which enables the systematic investigation of the distance distribution for various measures very swiftly. Recent research has shown that for practical purposes, MD plots outperform box plots, violin plots, and bean plots in several experiments with estimating the PDFs of variables of artificial and natural datasets [Thrun et al., 2020].

The PDE approach for estimating the *PDF* of $df$ is especially focused on finding multimodal distributions. Let *PDF* be a distribution mixture of weighted distributions F with weights $w_i$ depending on parameters $S_i$; then,

$$PDF(df) = \sum_{i=0}^{M} w_i \, F(df|\{S_i\}) = \sum_{i=1}^{M} p(c_i) \cdot p(df|c_i) \qquad (1)$$

where the likelihood of generating data in a component of the mixture is the conditional probability $p(df|c_i)$ and the probability of choosing a class $c_i$ is the prior $p(c_i)$ of mode $i$ for two or more modes M. In the particular case of a Gaussian mixture model (GMM) with means $m_i$ and standard deviations $s_i$, Eq. 1 yields

$$PDF(df) = \sum_{i=0}^{M} w_i \, N(df|m_i, s_i) = \sum_{i=1}^{M} w_i \cdot \frac{1}{\sqrt{2\pi s_i^2}} \cdot e^{-\frac{(df-m_i)^2}{2s_i^2}} = \sum_{i=1}^{M} p(c_i) \cdot p(df|c_i) \qquad (2)$$

Usually, the parameters of a GMM can be modeled by the EM algorithm [Dempster et al., 1977]. For every distance feature $df$, Hartigan's dip test against uniform distributions is performed [Hartigan/Hartigan, 1985] within the MD plot. If the MD plot does not yield a unimodal distribution, Gaussian mixture modeling can be performed as in [Ultsch et al., 2015]. The assumption is that the intra-pd lie in the left mode(s), the inter-pd is in the right (last) mode and outliers (if any) are outside the left (first) mode.

In such a case, we can use Bayes' theorem (e.g., [Duda et al., 2001]) to estimate the boundaries between modes by calculating the posterior $p(c_i|df)$ as

$$p(c_i|df) = \frac{p(df|c_i)p(c_i)}{\sum_{i=1}^{M} p(df|c_i)p(c_i)} \qquad (3)$$

We usually set the Bayesian decision boundaries *BD* with $p(c_i|df) = 0.5$ to define an equal probability of being in two classes as the boundary.

## 2.4. Curse of dimensionality in clustering

The term "curse of dimensionality" was introduced by Bellmann [Bellman, 1961]. Beyer et al. (1999a) showed that as a direct consequence, under a broad range of practically relevant conditions, the difference between the distances of a randomly chosen point to its furthest and nearest neighbors decreases as the dimensionality increases.

Let $l_d \in I^d$ be a data point of dimensionality d, $\|.\|_k$ the distance ($L^k$ metric) of a vector to a fixed randomly chosen point, e.g., the origin (0,…,0), $E(.)$ be the expected value, $var(.)$ be the variance,

$Dmin = min(\|l_d\|_k)$ be the distance from the nearest neighbor and $Dmax = max(\|l_d\|_k)$ be the distance from the farthest neighbor to the origin (0,…,0), and $P$ the probability; then [Aggarwal et al., 2001] outlined based on the theorem 1 of [Beyer et al., 1999b] (equation 1 there) that,

$$\forall \epsilon > 0: \text{If } \lim_{d \to \infty} var\left(\frac{\|l_d\|_k}{E(\|l_d\|_k)}\right) = 0, then \lim_{d \to \infty} P\left(\frac{Dmax - Dmin}{Dmin} \leq \epsilon\right) = 1 \quad (4)$$

In words, under a broad range of practically relevant conditions, all points tend to be equally distant from each other in a high-dimensional space because the observed distance between any two points tends towards a constant as the dimension increases [Verleysen et al., 2003].

Combining the concept of intra-pd with the triangle inequality leads to the **second assumption**: the relative relationships between high-dimensional data points define natural partitions of data. This implicitly means that the curse of dimensionality (c.f. [Beyer et al., 1999a; Verleysen et al., 2003]) does not apply as along as the distance distribution is multimodal, because otherwise, as the dimensionality increases, the distance to the nearest point approaches the distance to the farthest point[Beyer et al., 1999a], which leads to a narrow unimodal distribution of distances. Hence, the investigation of the full distance distribution allows investigating the effect of Eq. 4.

## 2.5. Distance selection

Given an appropriately preprocessed dataset, the distance features $df$ are computed for a set of accessible (e.g., [Eckert, 2018]) or otherwise given distance metrics. Next, the appropriate choice of distance is identified by the human-in-the-loop through the MD plot based on multimodality. The MD plot internally combines statistical testing with a density estimation approach that is particularly suitable for identifying multimodal structures.[Thrun et al., 2020] After a distance metric with a multimodal distance distribution is selected, $df$ is modeled with a Gaussian mixture model (e.g.,[Ultsch et al., 2015]). With Bayes' theorem, the posteriors can be computed for Gaussian mixtures that yield Bayesian boundaries. Then, the distribution $f_{intra(t_p)}$ of the intrapartition distances per partition $t_p$ can be compared to the full distance distribution after clustering. If we assume in the 1st approximation that each intrapartition feature $df_{intra(t_p)}$ follows a Gaussian distribution $N\left(df_{intra(t_p)}\big|m,s\right)$ with a robustly estimated mean $m$ and robustly estimated standard deviation $s$, then the PDF of distribution $f_{intra(t_p)}$ of its intrapartition distances should lie below the Bayesian boundary $BD$ that divides intrapartition from interpartition distances in the Gaussian mixture model with

$$m\left(df_{intra(t_p)}\right) + 2 * sd\left(df_{intra(t_p)}\right) \leq BD \; \forall t_p \quad (5)$$

Eq. 5 states that at least 95% of intrapartition distances lie below the Bayesian boundary $BD$. Please note, that in the following tables the short renovation $m(t_p)$ and $sd(t_p)$ will be used.

Visually, MD plots allow the human-in-the-loop a more detailed look at every distribution's PDF of each partition's intrapartition distances $df_{intra(t_p)}$ and whether the distribution lies below the Bayesian boundary $BD$. Detailed descriptions of the datasets and clustering algorithms used in the results can be found in [Thrun/Ultsch, 2020a] and [Thrun/Stier, 2021].

## 3. Results

The results section is divided into four parts. The first part outlines the two assumptions behind this work. The second part presents two artificial datasets showing the difference between distance-based structures and relative relationships between points that are not distance-based. The third part shows an application to two natural datasets.

### 3.1. Archetype of spatial relationships between data points

In Fig. 2 (top), two clusters are simulated by drawing a sample of 250 points for each cluster from a Gaussian random number generator with a mean of 0 and variance of 0.1 for the first feature (y-axis) and a shift $s_x$ in the mean $m_x = 0 \pm s_x$ and a variance of 0.1 for the second feature. In three steps of 0.1 each, the clusters are increasingly separated by the $shift\ s_x$. This allows for a clear separation of the clusters by the human eye with increasing values of $s_x$. In the second and third steps, the clusters start to become separated by the relative relationship between the points, resulting in a bimodal distribution of the Euclidean distances if the values of $s_x$ are sufficiently high. The distributions of the distances and intra-pds per cluster are shown in the MD plot on the bottom of each scatter plot of two-dimensional data on the left. The full distribution can be modeled with a GMM of two modes, and statistical testing can be applied (Table 1). First, the p-value of Hartigan's dip test statistic for unimodality is presented. Next, Table 1 presents the p-values of the GMMs in the chi-square test of the R package 'AdaptGauss' on CRAN[Ultsch et al., 2015] to investigate whether there is a significant difference between the GMM of the distribution of the distances and the distribution of the distances itself. The Bayesian decision boundary at $p(c_i|df) = 0.5$ (Eq. 3) is between the intra- and inter-pd, and the average inter-pd lies above the Bayes decision boundary if a model can be generated using 'AdaptGauss' that is significant w.r.t. the chi-square test implemented there.[Ultsch et al., 2015]

Table 1. Results of the GMM obtained by Eq. 2 for Fig. 2. If a valid GMM can be found, then the intra-pd are smaller than most of the inter-pds. The Bayes decision and chi test were computed with the R package AdaptGauss available on CRAN. Abbr.: I: No. of intra-pd above the BD in percent, BD: Bayes decision boundary; m (inter-pd), median of the interpatition distances, $m\ (t_x)$: median of the intra-pd of partition(cluster) $df_{intra(t_x)}$; sd: its robustly estimated standard deviation.

| Shift $s_x$ | Dip test | Chi test | BD | I in % | m (inter-pd) | $m(df_{intra(t_1)})$ $+2sd(df_{intra(t_1)})$ | $m(df_{intra(t_2)})$ $+2sd(df_{intra(t_2)})$ |
|---|---|---|---|---|---|---|---|
| 0.1 | 1 | 0 | 0.3 | / | 0.24 | 0.17+0.2 | 0.17+0.2 |
| 0.2 | 1 | 0.99 | 0.36 | 5.1 | 0.42 | 0.17+0.2 | 0.17+0.2 |
| 0.3 | 0 | 0.97 | 0.43 | 1.8 | 0.6 | 0.17+0.2 | 0.17+0.2 |

### 3.2. Distance-based cluster structures

In this section, the distinction between distance-based cluster structures and non-distance-based cluster structures is outlined. In Fig. 3 left, the Golfball dataset of the fundamental clustering problems suite (FCPS)[Thrun/Ultsch, 2020a] is presented, and the intra-pd distance distribution defined by k-means clustering is presented in Fig. 4, left. The intra-pd distribution does not possess a smaller range than the full distance distribution, and no multimodality is given (Hartigan's dip test p-value=1). Although the dataset is clearly based on the relative relationship between the data points and the dataset can be partitioned by dividing the sphere into two parts, no distance-based structures exist. Fig. 3 (middle and right) presents the Atom dataset of the FCPS[Thrun/Ultsch, 2020a]. Using Euclidean distances (Fig. 3, middle), the distribution of the intra-pd for cluster two is as large as the distribution of the full distance matrix, meaning that the intra-pd is not smaller than the inter-pd (Fig. 3, middle). The full distance matrix has a distribution with two peaks, but Ward's[Ward Jr, 1963] clustering is unable to separate these clusters, yielding an accuracy of 66%, as shown in Fig. 3, middle. By changing the Cartesian metric space (x,y,z) to the spherical metric space $(r, \phi, \theta)$ with a distance definition based on the radius $D(l(r,\phi,\theta), j(r,\phi,\theta)) = |r_l - r_j|$, both clusters obtain small intra-pd distributions, and the full distance distribution becomes separated more clearly into two modes (Fig. 4, right). The structures are distance based if a correct metric space is found. Using such a distance, Ward clustering has an accuracy of 100%.

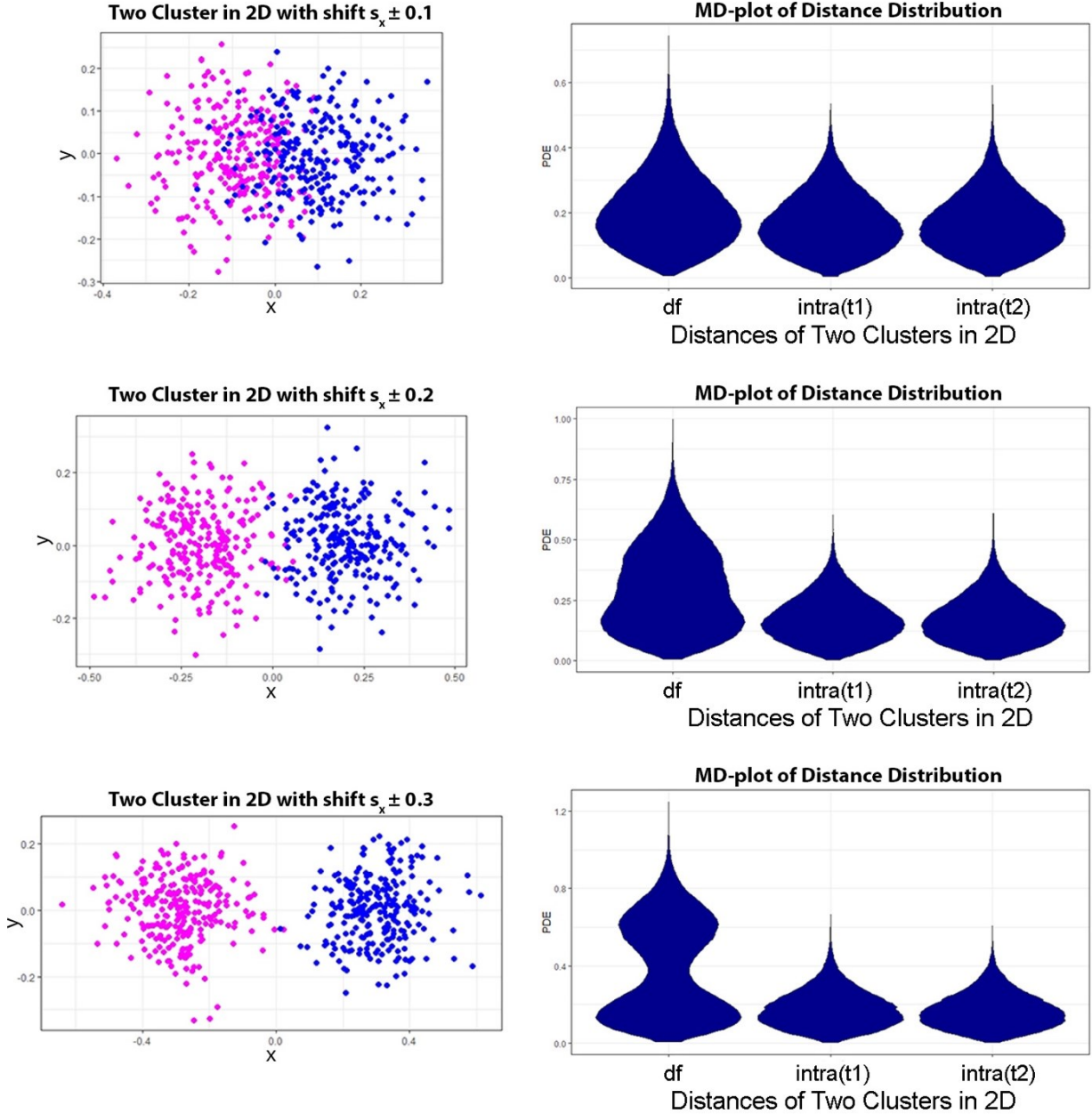

Fig. 2. Two clusters in 2D are shown on the left, with the MD plot (right) of the full distance distribution of df as well as the intra-pd distribution of $intra(t_x)$. The clusters are separated in three steps from top to bottom by shifting the cluster centers with $s_x$ in three steps resulting in a bimodal distance distribution and a smaller intra-cd distribution than the full distance distribution for increasing values of the shift $s_x$

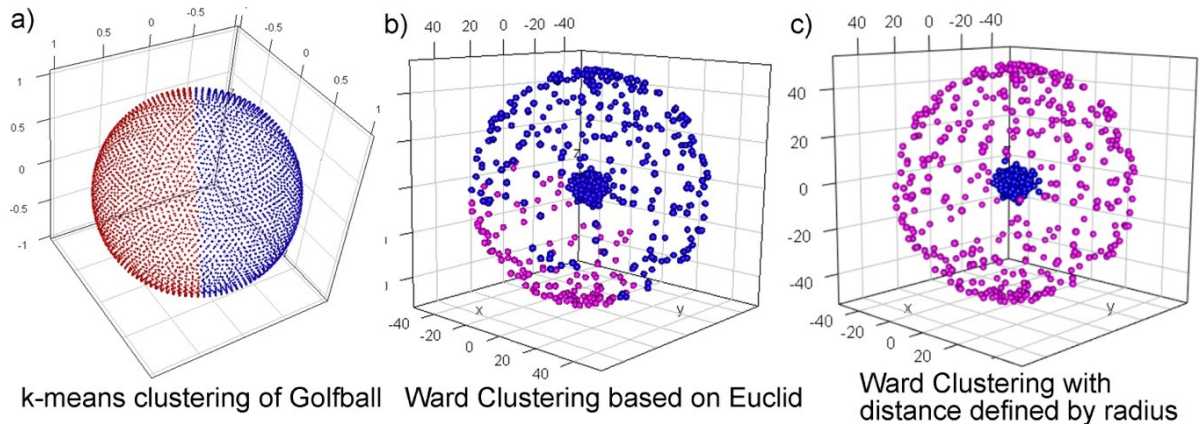

Fig. 3. <u>Left (a)</u>: The Golfball dataset of the FCPS [Thrun/Ultsch, 2020a] clustered with k-means, which is depicted by the colored points. <u>Middle (b)</u>: Ward's[Ward Jr, 1963] clustering of the Atom dataset of the FCPS[Thrun/Ultsch, 2020a] with a nonlinear distance-based structure. The correct structure is not reproduced by the clustering. <u>Right (c)</u>: Changing the metric space from Cartesian coordinates to spherical coordinates allows a distance to be defined based on the radius. By using this distance with Ward, the resulting clustering matches the prior classification of Atom (visualized in Cartesian coordinates). See Thrun/Ultsch (2020a) for a description of the data.

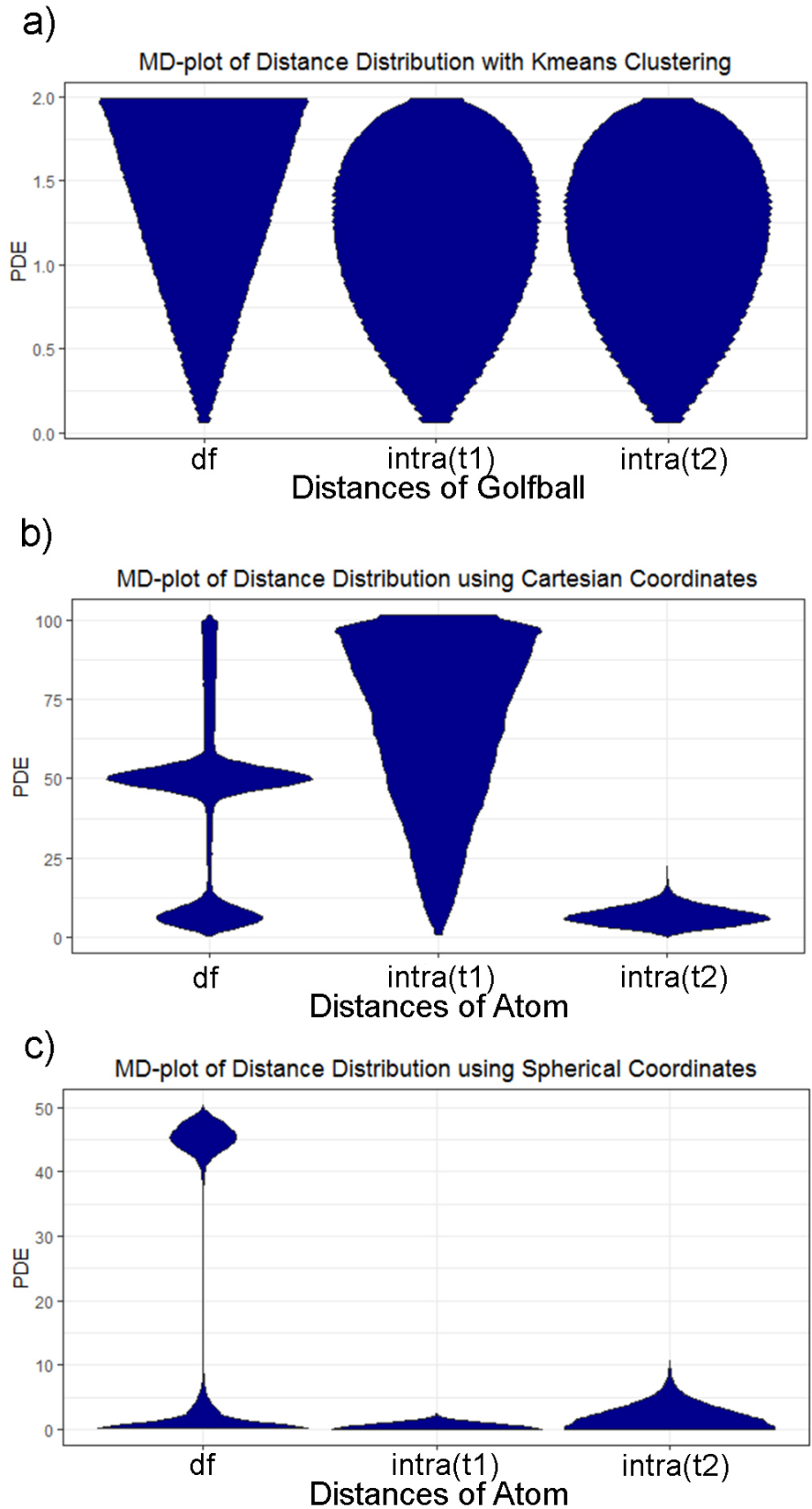

Fig. 4. Top (a): The Golfball dataset with the MD plot of its full Euclidean distance distribution and intra-pd defined by k-means clustering. Although the dataset is defined by the relative relationships between points (Fig. 3, left) and an appropriate partition is possible, there is no bimodality visible in the MD plot, and the intra-pd distributions have a large variance, similar to the full distance distribution. Middle (b): MD plot of Ward's clustering of Atom with its full Euclidean distance distribution and intra-pd. Bottom (c): For the case of a distance in spherical coordinates defined by the radius, the MD plot shows small intra -pd distributions for both clusters and two modes for the full distance distribution, which are clearly separated.

### 3.3. Exploration of distance distributions for the task of clustering

A stock dataset is investigated in the next example. The full distance distribution is not unimodal (Hartigan's dip test p<0.001) if chord distances are chosen (Fig. 5). The full distance distribution can be modeled approximately by two modes of the GMM (Fig. 5 left). The chi-square test yields p<0.01, but the QQ plot[Michael, 1983] indicates a good model (Fig. 5, right) with a resulting Bayesian boundary BD=0.79. In,[Thrun, 2019b] Databionic swarm clustering [Thrun/Ultsch, 2021] was applied to the chord distances, yielding a meaningful clustering, which was verified internally by the topographic map of the generalized U-matrix[Ultsch/Thrun, 2017; Thrun/Ultsch, 2020b] and externally by the heatmap.

Table 2 presents results for Eq. 5, for which conventional clustering algorithms are applied and the chord distance and intra-pd distances are computed. The intra-pd distances of the Databionic swarm clustering are below the Bayesian boundary BD.

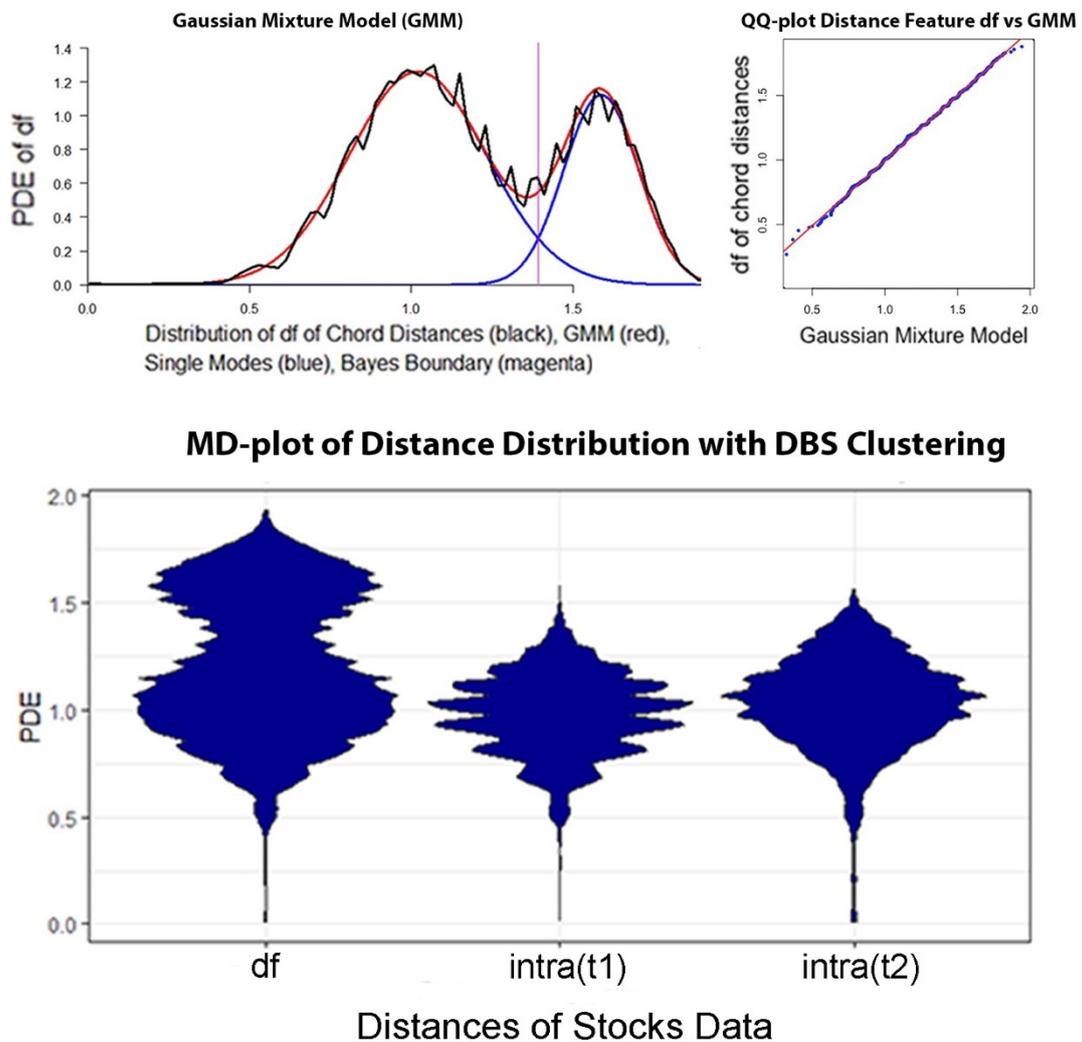

Fig. 5. Top left and right: GMM and QQ plot of the GMM versus the chord distance distribution of df. A straight line can be fitted, indicating a good model. Bottom: the MD-plot of the chord distances of the high-dimensional stock dataset shows two modes in the full distance distribution and a variance of intra-pd distributions that does not reach the second mode of the full distance distribution.

Table 2. Results of the distance-based clustering algorithms available in the R package "FCPS" on CRAN [Thrun/Stier, 2021] in comparison to the published results of Databionic Swarm, which is the only algorithm for which the intra-partition distances are below the Bayesian boundary of $BD = 1.40$. Abbr.: $m(t_x)$: median Intra-pd $df_{intra(t_x)}$ of partition (cluster) X; sd: its robustly estimated standard deviation; NA: one data point in the cluster .

| Clustering Algorithm | $m(t_1) + 2sd(t_1)$ | $m(t_2) + 2sd(t_2)$ |
|---|---|---|
| Ward | 1.01+0.41 | 1.09+0.41 |
| SingleL | 1.15+0.55 | NA |
| CompleteL | 1.01+0.40 | 1.10+0.40 |
| AverageL | 1.01+0.41 | 1.09+0.41 |
| WPGMA | 1.01+0.40 | 1.10+0.40 |
| MedianL | 1.15+0.55 | NA |
| CentroidL | 1.15+0.54 | NA |
| Minimax | 1.01+0.40 | 1.09+0.40 |
| MinEnergy | 1.01+0.41 | 1.09+0.41 |
| Gini | 1.01+0.39 | 1.11+0.30 |
| HDBSCAN | 1.15+0.55 | NA |
| Databionic Swarm | 1.00+0.37 | 1.03+0.37 |

The distribution of the full distance feature df and the distributions of the intra-pd are presented in Fig. 5, bottom for the chord distance. The variance of the distribution is in the range of the first mode of the full distance distribution.

For the Tetragonula dataset [Franck et al., 2004], the shared genetic allele distance measure is selected (see SI A for details), and the distance is not unimodal (in Hartigan's dip test, p<0.001). The distance distribution can be modeled by a GMM with three modes, in which the first and second modes represent the intra-pd and the third mode represents the inter-pd (Fig. 6 left). The chi-square test yields p<0.01, but the QQ plot indicates a good model (Fig. 6, right). Table 3 presents the results for Eq. 5, for which conventional clustering algorithms are applied with the chord distance and then the intra-pd distances are computed. The intra-pd distances of the average linkage clustering are below the Bayesian boundary BD. The WPGMA algorithm provides the same clustering, and hence, the same results for Eq. 5. Fig. 6, bottom, shows the MD plot of the average linkage clustering [Hennig, 2014], where the extracted distance-base structure was meaningful to the domain expert and was externally evaluated by several indices. The small variance of the intra-pd distribution relative to the large variance of the full distance distribution is visible for all clusters other than cluster 5. The ordering of the clusters in Fig. 6 is by descending cluster size and is the same as in Table 3.

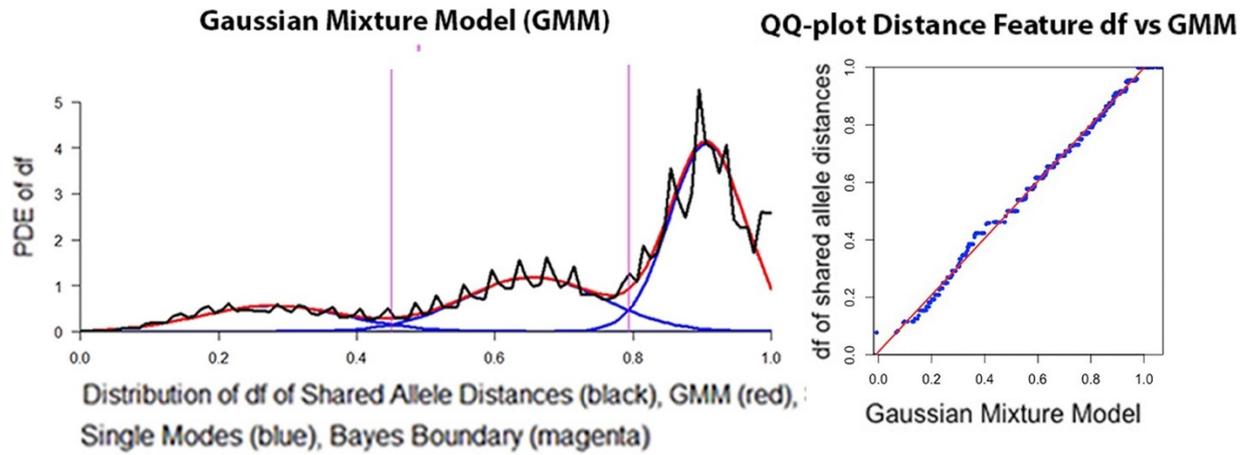

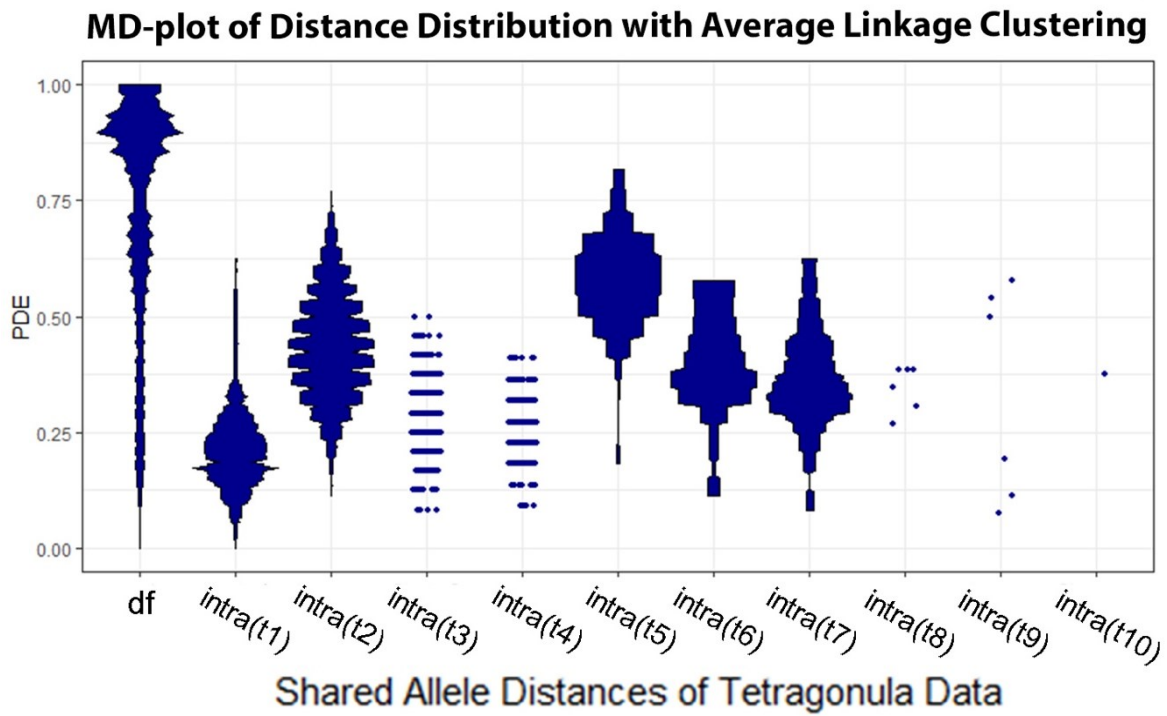

Fig. 6. Top left and right: GMM and QQ plot of the GMM versus the shared allele distance distribution of df. A straight line can be fitted, indicating a good model. Bottom: Average linkage clustering of the Tetragonula dataset[Hennig, 2014] with shared allele distance[Hausdorf/Hennig, 2010] results in 10 clusters. The MD plot visualizes the full distance distribution of df on the left and for each cluster the intrapartition distances as desrcibed in section 2.5. Cluster 5 can summarize the outliers.

Table 3. Results of the distance-based clustering algorithms available in the R package "FCPS" on CRAN[Thrun/Stier, 2021] in comparison to the published results of average linkage, which is an algorithm for which the intra-partition distances are below the Bayes boundary of $BD = 0.79$. The bolded values are above the BD. Abbr.: $m(t_x)$: median of Intra-pd $df_{intra(t_x)}$ of partition (cluster) x; sd: its robustly estimated standard deviation; NA: one data point in the cluster.

| $m(t_x) + 2sd(t_x)$ | Ward | SingleL | CompleteL | AverageL | WPGMA | MedianL | CentroidL | Minimax | MinEnergy | Gini | HDBSCAN |
|---|---|---|---|---|---|---|---|---|---|---|---|
| 1 | 0.21 +0.18 | **0.58 +0.40** | 0.27 +0.36 | 0.21 +0.17 | 0.21 +0.17 | **0.69 +0.50** | **0.58 +0.38** | 0.33 +0.36 | 0.21 +0.18 | 0.22 +0.18 | **0.58 +0.38** |
| 2 | 0.38 +0.12 | 0.29 +0.12 | 0.38 +0.12 | 0.42 +0.22 | 0.42 +0.22 | 0.29 +0.12 | **0.59 +0.42** | 0.40 +0.18 | 0.38 +0.18 | 0.48 +0.28 | **0.59 +0.44** |
| 3 | 0.21 +0.12 | 0.58 +0.14 | **0.55 +0.41** | 0.21 +0.12 | 0.21 +0.12 | 0.69 +0.10 | 0.58 +0.12 | 0.29 +0.12 | 0.21 +0.12 | 0.29 +0.14 | 0.58 +0.12 |
| 4 | 0.38 +0.14 | 0.29 +0.20 | 0.38 +0.12 | 0.42 +0.13 | 0.42 +0.13 | 0.29 +0.12 | 0.59 +0.14 | 0.26 +0.14 | 0.38 +0.14 | 0.28 +0.12 | 0.59 +0.14 |
| 5 | 0.29 +0.20 | 0.27 +0.14 | 0.55 +0.18 | 0.29 +0.15 | 0.29 +0.15 | 0.37 +0.00 | 0.29 +0.10 | 0.57 +0.20 | 0.29 +0.20 | 0.58 +0.18 | 0.29 +0.10 |
| 6 | 0.27 +0.16 | 0.55 +0.10 | 0.29 +0.14 | 0.27 +0.17 | 0.27 +0.17 | 0.12 +0.00 | 0.33 +0.12 | **0.48 +0.38** | 0.27 +0.16 | **0.54 +0.46** | 0.33 +0.01 |
| 7 | **0.59 +0.18** | 0.33 +0.01 | 0.38 +0.06 | 0.59 +0.15 | 0.59 +0.15 | 0.38 +0.00 | 0.37 +0.00 | 0.40 +0.18 | **0.59 +0.18** | 0.32 +0.16 | 0.37 +0.00 |
| 8 | 0.33 +0.14 | NA | 0.33 +0.10 | 0.38 +0.10 | 0.38 +0.10 | NA | 0.12 +0.00 | 0.42 +0.14 | 0.33 +0.14 | 0.28 +0.18 | NA |
| 9 | 0.38 +0.10 | NA | **0.50 +0.46** | 0.33 +0.46 | 0.33 +0.46 | NA | 0.38 +0.00 | 0.54 +0.35 | 0.38 +0.10 | 0.40 +0.18 | NA |
| 10 | **0.33 +0.46** | NA | 0.37 +0.01 | 0.37 +0.00 | 0.37 +0.00 | NA | NA | 0.38 +0.00 | 0.33 +0.12 | 0.31 +0.12 | NA |

## 4. Discussion

Two assumptions are made in this work. First, a clustering of distance-based structures is appropriate if it results in partitions with small distances within partitions of the data and large distances between partitions of the data (c.f. [Bouveyron et al., 2012]). Consequently, the second assumption is that the relative relationships between high-dimensional data points are the vital distinctions between clusters. Then, the differences between these relationships are strong enough to be visible in the distance distribution (see Figs. 2 and 4), which becomes non-unimodal. These two assumptions lead to the modeling of distance distributions with Gaussian mixture models (GMMs). Using the toolset for modeling GMMs given in [Ultsch et al., 2015] and mirrored-density plots (MD plots),[Thrun et al., 2020] the hypothesis of multimodality is tested on three artificial datasets as well as two high-dimensional datasets for the task of clustering. The first example presented in Fig. 1 shows that the more two clusters become separated, the more bimodal the distance distribution becomes and the better it can be modeled with a GMM of two modes. A similar artificial dataset was investigated in [Bozkaya/Ozsoyoglu, 1997]. However, the two clusters were either not sufficiently well separated or their density estimation approach was not sensitive enough to discover structures in continuous data which lead the authors to the assumption that the Euclidean distance was not an appropriate choice for their specific problem.

The next example shows that a distance-based cluster structure is visible through a nonunimodal distance distribution and finds that the right metric space changes the distance distribution to a clear bimodal distribution with intra-pd distributions that do not overlap with the last mode of the full distance distributions (Figs. 3 and 4). The assumption here is that the last mode of the full distance distribution represents the inter-pd. The last artificial example of Golfball clearly restricts the investigation in this paper (Figs. 3 and 4). Datasets defined by the relative relationships between points may be partitioned even if no distance-based structures exist. In such a case, the distance distributions will be unimodal, and the intra-pd distribution will have a large variance. Whether such a partition is meaningful has to be discussed by domain experts based on the data context. Here, it is argued that in such cases, a distance measure has to be defined w.r.t. the semantic meaning behind the data.

Nevertheless, if the detection of distance-based structures is stated as the goal of cluster analysis, then one can systematically investigate distance measures for the existence of multimodal distance distributions that can be used for the task of clustering. This is shown in the first example of the stock data, where the chosen distance measure was bimodal, resulting in meaningful clustering (Fig. 5). Using the Bayesian boundary of a GMM of the distance distribution, it was shown that in terms of intrapartition distances, the published clustering outperforms the results of 11 comparable clustering algorithms because for the compared clustering, the size of the intrapartition distances was higher than the Bayesian boundary.

Moreover, for context-based distance metrics, the distribution can still be exploited to investigate the clustering solution, which is shown in the example of the Tetragonula dataset (Fig. 6). Often, one dataset has more than one optimal partition (as is the case here; see the Tetragonula clustering in [Thrun/Ultsch, 2021]). Then, an appropriate choice would be to select the partition in a data-driven way where the intra-pd is nearer to zero, has a smaller variance and is below the Bayes boundary of the GMM if no additional knowledge about the data or a specific goal is available. Such a choice would be strictly data-driven and unbiased, which is contrary to unsupervised or supervised quality measures that have biases (c.f. the discussion in [Handl et al., 2005; Ball/Geyer-Schulz, 2018] and thus requires the need to account for their biases. For example, the published result [Hennig, 2014] outperforms the comparable algorithms in terms of intracluster distance distributions if the number of clusters remains fixed.

In sum, the approach presented here provides a first impression of which data-driven properties a valid distance measure should possess. This work proposes an exploratory investigation of distance measures trough distribution analysis during the process of modeling data before the conventional evaluation steps, especially if no prior knowledge about the data is available. To date, this critical step in modeling remains largely ignored.

## 5. Conclusion

This work is the first step in a data-driven exploration of a distance measure independent of the underlying context of the data. The results indicate that searching for multimodality in distance distributions can be reasonable if no prior knowledge about the data is available. This approach provides a data-driven choice for the selection of an appropriate distance measure and enables the evaluation of clustering solutions using Gaussian mixture models (GMMs) of the selected distance distributions under the assumption that distance-based structures are sought. Multimodality can be investigated with a mirrored-density plot (MD plot). The MD plot combines statistical testing for unimodality with a PDF estimation procedure focused on multimodality with an elaborate visualization approach, enabling a specific search for a distance with a non-unimodal distribution. In R, this search can be performed easily through a combination of all the distance measures available in the R package "parallelDist" on CRAN

[Eckert, 2018], which allows the investigation of a large number of distance measures with a short computation time due to the parallelization performed in C++.